\documentclass[letterpaper, 10 pt, conference]{ieeeconf}  

\IEEEoverridecommandlockouts                              

\usepackage{graphics} 
\usepackage{epsfig} 
\usepackage{mathptmx} 
\usepackage{amsmath} 
\usepackage{amssymb}  

\usepackage{mathtools}
\usepackage{algorithm}
\usepackage{algpseudocode} 
\usepackage{multirow}
\usepackage{booktabs}
\usepackage{makecell}
\usepackage{placeins}

\title{\LARGE \bf
CLIS-MPPI: Closed-Loop Model Predictive Path Integral Control With Initial-State Sampling Under Execution Delay
}

\author{Zidong Liu$^{1}$ and Xu Chen$^{\dagger, 1}$
\thanks{$^\dagger$: corresponding author.}
\thanks{$^{1}$Department of Mechanical Engineering, University of Washington,  Seattle, WA, USA. {\tt\small\{liuzd, chx\}@uw.edu}}%
}

\begin{document}

\maketitle
\thispagestyle{empty}
\pagestyle{empty}

\begin{abstract}

Sampling-based motion planning such as Model Predictive Path Integral (MPPI) control has become attractive for robotics due to its ability to handle nonlinear dynamics and non-smooth objectives without gradient information, while leveraging parallel rollout computation for real-time performance. However, in real-world robotic systems, the optimized control input is often applied after a non-negligible planning delay, causing a mismatch between the state used in forward simulation and the actual state at control execution, which can significantly degrade planning accuracy and may even lead to unsafe behavior such as collisions. This paper proposes Closed-Loop MPPI with Initial-State Sampling (CLIS-MPPI), which explicitly addresses execution-time state mismatch by modeling planning delay as a distribution over plausible execution-time initial states, performing rollout optimization from sampled initial states, and applying a rollout-based local feedback correction during control execution.
The proposed method is validated through real-world experiments on a 6-DoF UR5e manipulator in a narrow-passage task involving obstacle avoidance and end-effector target reaching. 
Across 5 real-world runs of the same task, standard MPPI collided in 2 runs, whereas CLIS-MPPI remained collision-free in all 5 runs. Averaged over all collision-free runs for each method (3 for MPPI and 5 for CLIS-MPPI), CLIS-MPPI reduced the average task completion time from 25.4~s to 24.4~s and the average end-effector path length from 0.456~m to 0.425~m. It also reduced the overall joint-position and joint-velocity root mean square errors (RMSEs) from 0.100~rad and 0.108~rad/s to 0.017~rad and 0.011~rad/s, respectively. These results demonstrate improved safety, planning-execution consistency, and motion smoothness.

\end{abstract}

\section{INTRODUCTION}

With the rapid growth of computing hardware, sampling-based motion planning methods~\cite{kazim2024recent,orthey2023sampling}, especially Model Predictive Path Integral (MPPI) control~\cite{williams2018information}, have emerged as a powerful framework for real-time robotic motion planning and control. MPPI repeatedly samples stochastic control rollouts, performs forward simulation, evaluates trajectory costs, and refines the control sequence through a path-integral weighted update over a receding horizon. Compared with traditional gradient-based model predictive control (MPC)~\cite{schwenzer2021review}, MPPI does not require differentiable cost functions, making it particularly suitable for robotic systems with nonlinear dynamics, obstacle-rich environments, and complex non-smooth objectives~\cite{bhardwaj2022storm}. Furthermore, unlike learning-based methods such as reinforcement learning~\cite{tang2025deep}, which typically depend on large-scale offline training, MPPI is a fully online planning method and can leverage GPU-parallelized rollout computation~\cite{lee2024gpu} to enable real-time decision-making.

\begin{figure}[htbp]
    \centering
    \includegraphics[width=1.0\linewidth]{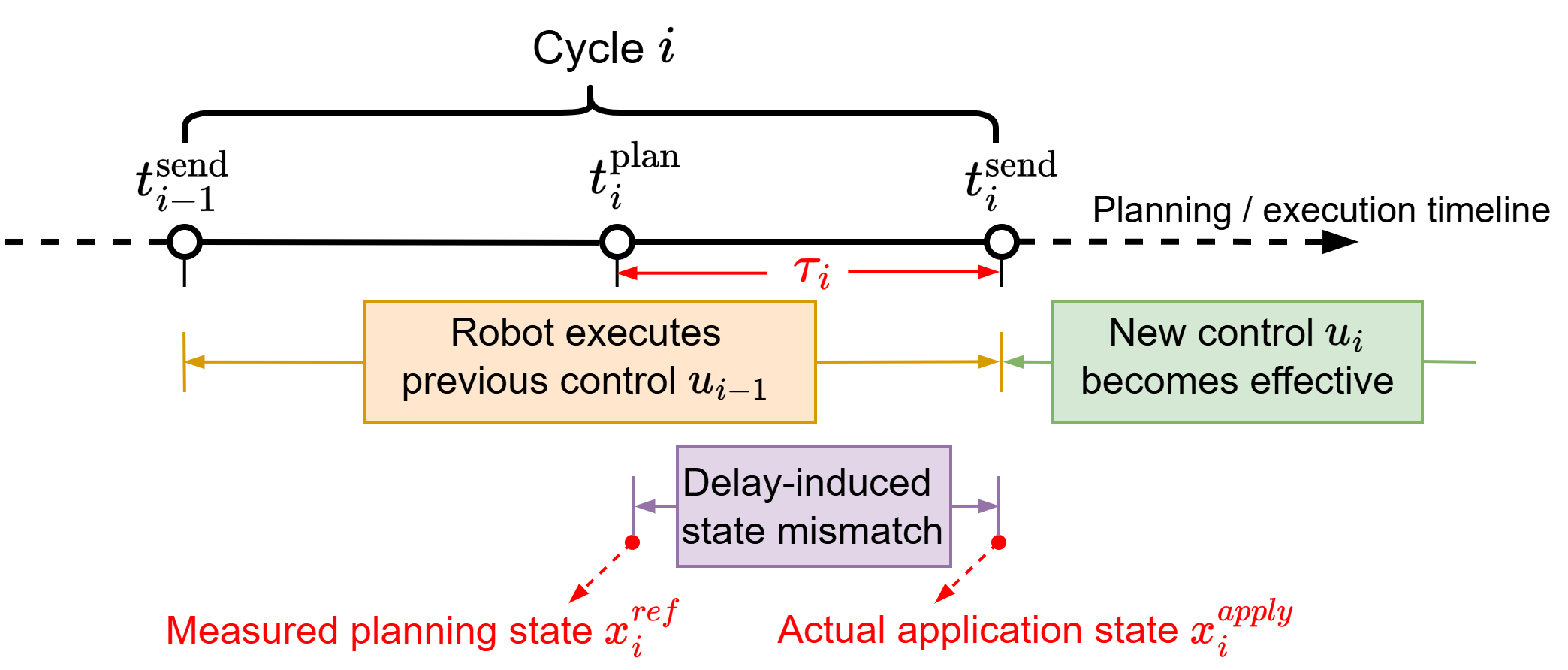}
    \caption{Illustration of execution-time state mismatch in the $i$-th planning cycle. At planning time $t_i^{\mathrm{plan}}$, MPPI optimizes using the measured state $x_i^{\mathrm{ref}}$. During the delay window, the robot continues executing the previous control $u_{i-1}$. Consequently, when the newly planned control $u_{i}$ actually becomes effective at $t_i^{\mathrm{send}}$, the system state has already evolved to $x_i^{\mathrm{apply}} \neq x_i^{\mathrm{ref}}$.}
    \label{delay}
\end{figure}

Despite these advantages, the performance of MPPI in real-world robotic systems can be significantly degraded by execution-time delay. In particular, the quality of MPPI critically depends on the accuracy of its forward simulation and cost evaluation, since both are used to determine the control update at each planning cycle. However, forward simulation is typically the most computationally expensive step in MPPI, and the resulting planning latency can introduce a substantial mismatch between the state used for optimization and the state at which the optimized control is actually applied. As illustrated in Fig.~\ref{delay}, at planning time $t_i^{\mathrm{plan}}$, MPPI starts forward simulation from the measured robot state $x_i^{\mathrm{ref}}$, and the rollout costs are evaluated based on this state. During the planning interval $\tau_i$, however, the robot continues executing the previous control command $u_{i-1}$. Consequently, when the newly optimized control $u_i$ becomes effective at $t_i^{\mathrm{send}}$, the actual robot state has already evolved to $x_i^{\mathrm{apply}} \neq x_i^{\mathrm{ref}}$. As a result, the control command $u_i$ is no longer applied to the state on which it was optimized, leading to a delay-induced state mismatch. This issue becomes particularly critical in high-frequency control settings or when the robot operates with large control actions, where even a small planning delay may produce a significant state deviation. In safety-critical tasks such as obstacle avoidance in narrow spaces, such mismatch can cause severe performance degradation and even collisions, despite the fact that the planned rollouts themselves are collision-free during forward simulation.

To address the delay-induced state mismatch, we propose closed-loop MPPI with initial-state sampling (CLIS-MPPI), as illustrated in Fig.~\ref{Framework}. The key idea is to explicitly convert execution delay into an initial-state uncertainty at each planning cycle, and to compensate for the resulting mismatch through a closed-loop control update. Instead of assuming that the control optimization starts from a single measured state, CLIS-MPPI samples a distribution of perturbed initial states around the currently measured state to represent the possible system evolution during the planning interval. For each sampled initial state, perturbed control sequences are generated and parallel forward simulations are performed to obtain rollout trajectories and their associated costs. Based on the weighted rollout distribution, a nominal state-control trajectory is constructed, from which a local feedback gain is further extracted in a rollout-based manner. The final control input is then computed as a closed-loop combination of the MPPI feedforward update and a state-feedback correction term. In this way, the proposed method enables the applied control command to better match the actual robot state at execution time, thereby improving consistency between planning and execution, motion smoothness, and safety in delay-sensitive real-world tasks.

In contrast to standard MPPI, which assumes immediate actuation after planning and samples only control perturbations from a single planning state, CLIS-MPPI replaces this assumption with a distribution over plausible execution-time initial states, jointly samples both initial-state and control perturbations, and augments the control update with rollout-based feedback correction.

\subsection{Related Work}
MPPI was originally introduced for aggressive autonomous driving tasks, where its sampling-based formulation demonstrated strong capability in handling high-speed, nonlinear systems under complex objectives~\cite{williams2018information}. Since then, MPPI has been extended to a wide variety of robotic platforms, including robotic manipulators~\cite{bhardwaj2022storm,vasilopoulos2023ramp,zhou2025parallel}, aerial robots~\cite{zhai2026pa}, and legged systems~\cite{alvarez2025real}. In robotic manipulation, STORM~\cite{bhardwaj2022storm} and RAMP~\cite{vasilopoulos2023ramp} showed that GPU-accelerated MPPI can achieve real-time obstacle avoidance and goal-reaching on 6--7 Degrees of Freedom (DoF) manipulators in cluttered environments. For dynamic environments, Zhou \emph{et al.}~\cite{zhou2025parallel} proposed a parallel MPPI framework for manipulators that improves computational efficiency and incorporates a gradient-velocity modulated signed distance field cost for dynamic obstacle avoidance. Zhai \emph{et al.}~\cite{zhai2026pa} developed a perception-aware MPPI method for quadrotor navigation in unknown environments, where perception-driven objectives are integrated into the planning process to promote exploration and path recovery under occlusion. 

Delay-induced state mismatch causes the optimized control input to be applied to a state that is no longer consistent with the one used during planning, which may lead to degraded performance and even unsafe behavior. To improve the safety of MPPI, several works have combined MPPI with control barrier functions (CBFs)~\cite{tao2022path,parwana2024model,rabiee2025guaranteed}. Tao \emph{et al.}~\cite{tao2022path} proposed SCBF-MPPI, which constrains the sampling distribution using stochastic control barrier function conditions to improve probabilistic safety and sample efficiency. Parwana \emph{et al.}~\cite{parwana2024model} combined an MPPI planner with a CBF-based safety filter so that MPPI handles task planning while the CBF layer enforces safe execution. Rabiee and Hoagg~\cite{rabiee2025guaranteed} proposed GS-MPPI, which incorporates a composite CBF into the system dynamics and enables MPPI to optimize only over guaranteed-safe trajectories. In a different direction, Belvedere \emph{et al.}~\cite{belvedere2025feedback} proposed Feedback-MPPI, which augments MPPI with local feedback gains extracted from rollout sensitivity analysis to improve closed-loop responsiveness. However, these methods mainly focus on safety filtering or high-frequency feedback enhancement, and do not explicitly address the execution-time state mismatch caused by planning delay.

\subsection{Contributions}
The main contributions of this paper are as follows:
\begin{enumerate}
    \item We propose CLIS-MPPI, a novel MPPI formulation that explicitly compensates for execution-time state mismatch. Unlike standard MPPI, which implicitly assumes immediate actuation after the planning stage, CLIS-MPPI models planning delay as a distribution over plausible execution-time initial states, jointly samples both initial-state and control perturbations, and computes the applied control input using a closed-loop combination of the MPPI feedforward update and a rollout-based local feedback correction.
   
    \item We validate CLIS-MPPI through real-world experiments on a UR5e manipulator in a narrow-passage task involving both obstacle avoidance and end-effector target reaching. Across 5 real-world runs of the same task, standard MPPI collides with the environment in 2 runs, whereas CLIS-MPPI remains collision-free in all trials. Averaged over all collision-free runs for each method (3 for MPPI and 5 for CLIS-MPPI), CLIS-MPPI reduced the average task completion time from 25.4~s to 24.4~s and shortened the average end-effector path length from 0.456~m to 0.425~m. In addition, the overall joint-position and joint-velocity root mean square errors (RMSEs) were reduced from 0.100~rad and 0.108~rad/s under MPPI to 0.017~rad and 0.011~rad/s under CLIS-MPPI, corresponding to reductions by factors of approximately 5.9 and 9.7, respectively.
\end{enumerate}

\section{Proposed Closed-Loop MPPI With Initial-State Sampling}

\begin{figure*}[t]
    \centering
    \includegraphics[width=1\textwidth]{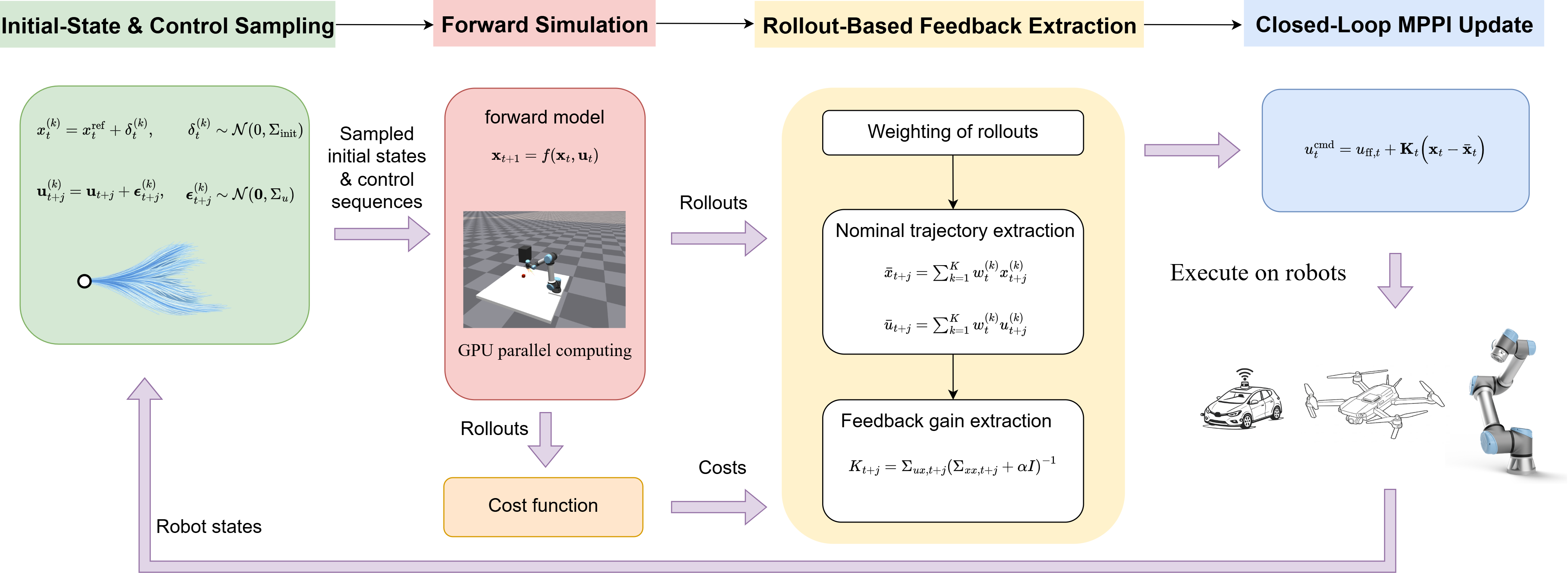}
    \caption{Overview of the proposed CLIS-MPPI. CLIS-MPPI converts execution delay into an initial-state uncertainty distribution, samples both perturbed initial states and control sequences, and performs parallel forward simulation to obtain rollout trajectories and costs. The weighted rollout distribution is then used to construct a nominal trajectory and extract local feedback gains. Finally, the applied control is computed as a closed-loop combination of the MPPI feedforward term and a state-feedback correction.}
    \label{Framework}
\end{figure*}

\subsection{MPPI Preliminaries}
\label{subsec:mppi-preliminaries}

We consider a general discrete-time nonlinear system
\begin{equation}
\mathbf{x}_{t+1} = f(\mathbf{x}_t,\mathbf{u}_t),
\end{equation}
where $\mathbf{x}_t \in \mathbb{R}^{n}$ and $\mathbf{u}_t \in \mathbb{R}^{m}$ denote the system state and control input, respectively. At each iteration $t$, MPPI optimizes a horizon-$N$ control sequence
\begin{equation}
\mathbf{U}_t \coloneqq \{\mathbf{u}_{t+j}\}_{j=0}^{N-1},
\end{equation}
where $j\in\{0,\dots,N-1\}$ indexes the within-horizon time step.

To evaluate the control sequence, MPPI generates $K$ stochastic rollouts by perturbing the nominal controls. For rollout index $k\in\{1,\dots,K\}$, Gaussian perturbations
\begin{equation}
\boldsymbol{\epsilon}^{(k)}_{t+j}\sim\mathcal{N}(\mathbf{0},\Sigma_u)
\end{equation}
are sampled and added to the nominal controls, yielding
\begin{equation}
\mathbf{u}^{(k)}_{t+j}
=
\mathbf{u}_{t+j}+\boldsymbol{\epsilon}^{(k)}_{t+j},
\qquad j=0,\dots,N-1.
\end{equation}
Starting from $\mathbf{x}^{(k)}_t=\mathbf{x}_t$, each rollout is propagated according to
\begin{equation}
\mathbf{x}^{(k)}_{t+j+1}
=
f\!\left(\mathbf{x}^{(k)}_{t+j},\mathbf{u}^{(k)}_{t+j}\right),
\qquad j=0,\dots,N-1.
\label{eq:mppi_rollout}
\end{equation}

Each rollout is assigned a trajectory cost
\begin{equation}
S_t^{(k)}
=
\phi\!\left(\mathbf{x}^{(k)}_{t+N}\right)
+
\sum_{j=0}^{N-1}
\ell\!\left(\mathbf{x}^{(k)}_{t+j},\mathbf{u}^{(k)}_{t+j}\right),
\label{eq:mppi_cost}
\end{equation}
where $\phi(\cdot)$ and $\ell(\cdot)$ denote the terminal and stage cost, respectively. The importance weight of rollout $k$ is then defined as
\begin{equation}
w_t^{(k)}
=
\frac{\exp\!\left(-\frac{1}{\lambda}S_t^{(k)}\right)}
{\sum_{r=1}^{K}\exp\!\left(-\frac{1}{\lambda}S_t^{(r)}\right)},
\label{eq:mppi_weight}
\end{equation}
where $\lambda>0$ is the temperature parameter.

Finally, MPPI updates the nominal controls \(\mathbf{u}_{t+j}\) via adding an importance weighted average of perturbations:
\begin{equation}
\mathbf{u}_{t+j}
\coloneqq
\mathbf{u}_{t+j}
+
\sum_{k=1}^{K} w_t^{(k)}\,\boldsymbol{\epsilon}^{(k)}_{t+j},
\qquad j=0,\dots,N-1.
\label{eq:mppi_update}
\end{equation}
Only the first control $\mathbf{u}_{t}$ is executed, and the procedure repeats in a receding-horizon manner.

Standard MPPI implicitly assumes that the optimized control can be applied immediately after planning and that the measured planning state coincides with the actual execution-time state. In real-world multi-body robotic systems, however, computation, communication, and scheduling delays violate this assumption, causing the control update to become stale at the moment of execution.

\subsection{Proposed CLIS-MPPI}
\label{subsec:clis-mppi}

We propose CLIS-MPPI, which augments standard MPPI through three components: 1) delay-induced initial-state uncertainty modeling, 2) initial-state-sampled MPPI rollouts, and 3) rollout-based local feedback extraction for closed-loop execution. Fig.~\ref{Framework} illustrates the overall workflow of the proposed CLIS-MPPI, and Alg.~\ref{alg:clismppi} provides the corresponding pseudocode.

\begin{algorithm}[t]
\caption{Proposed CLIS-MPPI}
\label{alg:clismppi}
\begin{algorithmic}[1]
\Require $\mathbf{x}^{\mathrm{ref}}_t$, $\mathbf{U}_t=\{\mathbf{u}_{t+j}\}_{j=0}^{N-1}$, $K,N,\lambda,\Sigma_u,\Sigma_{\mathrm{init}}$, $f(\cdot)$, $\ell(\cdot),\phi(\cdot)$, $\alpha$
\Ensure updated $\mathbf{U}_t$, $\{\mathbf{u}_{\mathrm{ff},t+j}\}_{j=0}^{N-1}$, $\{\bar{\mathbf{x}}_{t+j}\}_{j=0}^{N}$, $\{\mathbf{K}_{t+j}\}_{j=0}^{N-1}$

\State \textbf{[Rollouts]} \textbf{for each} $k=1,\dots,K$ \textbf{in parallel do}
\State \hspace{0.8em} sample $\boldsymbol{\delta}^{(k)}_t \sim \mathcal{N}(\mathbf{0},\Sigma_{\mathrm{init}})$ and set $\mathbf{x}^{(k)}_t \gets \mathbf{x}^{\mathrm{ref}}_t + \boldsymbol{\delta}^{(k)}_t$
\State \hspace{0.8em} \textbf{for} $j=0,\dots,N-1$ \textbf{do}
\State \hspace{1.6em} sample $\boldsymbol{\epsilon}^{(k)}_{t+j}\sim\mathcal{N}(\mathbf{0},\Sigma_u)$
\State \hspace{1.6em} set $\mathbf{u}^{(k)}_{t+j}\gets \mathbf{u}_{t+j}+\boldsymbol{\epsilon}^{(k)}_{t+j}$
\State \hspace{1.6em} propagate $\mathbf{x}^{(k)}_{t+j+1}\gets f(\mathbf{x}^{(k)}_{t+j},\mathbf{u}^{(k)}_{t+j})$
\State \hspace{0.8em} \textbf{end for}
\State \hspace{0.8em} compute $S_t^{(k)} \gets \phi(\mathbf{x}^{(k)}_{t+N})+\sum_{j=0}^{N-1}\ell(\mathbf{x}^{(k)}_{t+j},\mathbf{u}^{(k)}_{t+j})$
\State \textbf{end for}

\State \textbf{[Weighting]} compute $w_t^{(k)} \propto \exp(-S_t^{(k)}/\lambda)$ and normalize over $k=1,\dots,K$

\State \textbf{[Feedforward \& nominal trajectory]} \textbf{for} $j=0,\dots,N-1$ \textbf{do}
\State \hspace{0.8em} $\mathbf{u}_{\mathrm{ff},t+j}\gets \mathbf{u}_{t+j}+\sum_{k=1}^{K}w_t^{(k)}\boldsymbol{\epsilon}^{(k)}_{t+j}$
\State \hspace{0.8em} $\bar{\mathbf{x}}_{t+j}\gets \sum_{k=1}^{K}w_t^{(k)}\mathbf{x}^{(k)}_{t+j}$
\State \hspace{0.8em} $\bar{\mathbf{u}}_{t+j}\gets \sum_{k=1}^{K}w_t^{(k)}\mathbf{u}^{(k)}_{t+j}$
\State \textbf{end for}
\State \hspace{0.8em} $\bar{\mathbf{x}}_{t+N}\gets \sum_{k=1}^{K}w_t^{(k)}\mathbf{x}^{(k)}_{t+N}$

\State \textbf{[Feedback gain]} \textbf{for} $j=0,\dots,N-1$ \textbf{do}
\State \hspace{0.8em} compute centered deviations $\tilde{\mathbf{x}}^{(k)}_{t+j}$ and $\tilde{\mathbf{u}}^{(k)}_{t+j}$
\State \hspace{0.8em} $\Sigma_{xx,t+j}\gets \sum_{k=1}^{K} w_t^{(k)} \tilde{\mathbf{x}}^{(k)}_{t+j}\tilde{\mathbf{x}}^{(k)\top}_{t+j}$
\State \hspace{0.8em} $\Sigma_{ux,t+j}\gets \sum_{k=1}^{K} w_t^{(k)} \tilde{\mathbf{u}}^{(k)}_{t+j}\tilde{\mathbf{x}}^{(k)\top}_{t+j}$
\State \hspace{0.8em} $\mathbf{K}_{t+j}\gets \Sigma_{ux,t+j}(\Sigma_{xx,t+j}+\alpha\mathbf{I})^{-1}$
\State \textbf{end for}

\State \textbf{[Execute]} measure $\mathbf{x}_t$ and apply
\State \hspace{0.8em} $\mathbf{u}^{\mathrm{cmd}}_t\gets \mathbf{u}_{\mathrm{ff},t}+\mathbf{K}_t(\mathbf{x}_t-\bar{\mathbf{x}}_t)$

\State \textbf{[Shift]} warm-start the next horizon using $\{\mathbf{u}_{\mathrm{ff},t+j}\}_{j=1}^{N-1}$
\end{algorithmic}
\end{algorithm}

\subsubsection{Delay-Induced Initial-State Uncertainty}

Let $t$ denote the planning instant, and let the newly computed policy become effective only after an execution delay $\tau_t$. Denote by $\mathbf{x}^{\mathrm{ref}}_t$ the state used by the planner at time $t$, and by $\mathbf{x}^{\mathrm{apply}}_t$ the true system state when the new policy actually starts affecting the system. Their mismatch is defined as
\begin{equation}
\Delta \mathbf{x}_t
\triangleq
\mathbf{x}^{\mathrm{apply}}_t-\mathbf{x}^{\mathrm{ref}}_t.
\label{eq:state_mismatch}
\end{equation}
Instead of predicting a single delayed state, we model this execution-time mismatch as a random variable
\begin{equation}
\Delta \mathbf{x}_t \sim \mathcal{N}(\mathbf{0},\Sigma_{\mathrm{delay}}),
\label{eq:delay_cov}
\end{equation}
where $\Sigma_{\mathrm{delay}}$ captures the state uncertainty induced by execution delay, communication jitter, zero-order hold, and low-level tracking error.

For systems driven by streaming velocity commands, we model the dominant position mismatch as
\begin{equation}
\Delta q \approx \mathbf{u}_{\mathrm{hold}}\tau,
\end{equation}
where $\mathbf{u}_{\mathrm{hold}}$ is the average previously issued command during the delay window and \(\tau\) is the effective delay. Assuming approximate independence between $\mathbf{u}_{\mathrm{hold}}$ and \(\tau\), the position mismatch covariance can be approximated as
\begin{equation}
\Sigma_{\Delta q} \approx
\mathbb{E}[\tau^2]\Sigma_{u,\mathrm{hold}}+ \sigma_\tau^2\boldsymbol{\mu}_u\boldsymbol{\mu}_u^\top.
\end{equation}
where
\begin{equation}
\boldsymbol{\mu}_{u} = \mathbb{E}[\mathbf{u}_{\mathrm{hold}}],
\qquad
\Sigma_{u,\mathrm{hold}} = \mathrm{Cov}(\mathbf{u}_{\mathrm{hold}}),
\qquad
\sigma_\tau^2 = \mathrm{Var}(\tau).
\end{equation}
For a general state partitioned into position- and velocity-like components such as the case in robotics, we have
\begin{equation}
\Sigma_{\mathrm{delay}}
\approx
\mathrm{blkdiag}(\Sigma_{\Delta q},\Sigma_{\Delta \dot q}).
\label{eq:sigma_delay_blk}
\end{equation}
The covariance used for initial-state sampling is defined as
\begin{equation}
\Sigma_{\mathrm{init}}
=
\Sigma_{\mathrm{delay}}.
\label{eq:init_cov}
\end{equation}
The proposed method therefore directly uses the delay-induced execution-time mismatch covariance to sample plausible initial states for rollout propagation.

\subsubsection{Initial-State-Sampled MPPI}

Unlike standard MPPI, CLIS-MPPI samples both the execution-time initial state and the control sequence. For rollout $k$, we sample
\begin{equation}
\boldsymbol{\delta}^{(k)}_t \sim \mathcal{N}(\mathbf{0},\Sigma_{\mathrm{init}}),
\end{equation}
and initialize the rollout from
\begin{equation}
\mathbf{x}^{(k)}_t
=
\mathbf{x}^{\mathrm{ref}}_t + \boldsymbol{\delta}^{(k)}_t.
\label{eq:initial_state_sampling}
\end{equation}
The perturbed controls are sampled as
\begin{equation}
\mathbf{u}^{(k)}_{t+j}
=
\mathbf{u}_{t+j}+\boldsymbol{\epsilon}^{(k)}_{t+j},
\qquad
\boldsymbol{\epsilon}^{(k)}_{t+j}\sim\mathcal{N}(\mathbf{0},\Sigma_u),
\end{equation}
and the rollout is propagated according to
\begin{equation}
\mathbf{x}^{(k)}_{t+j+1}
=
f\!\left(\mathbf{x}^{(k)}_{t+j},\mathbf{u}^{(k)}_{t+j}\right),
\qquad j=0,\dots,N-1.
\label{eq:clis_rollout}
\end{equation}
Therefore, CLIS-MPPI optimizes the control sequence over a distribution of plausible execution-time initial conditions rather than a single assumed planning state.

\subsubsection{Rollout-Based Feedback Extraction and Closed-Loop Execution}

Beyond the feedforward MPPI update, CLIS-MPPI extracts a local state-feedback gain from the weighted rollout distribution. For each within-horizon step $j$, we first compute the weighted nominal state and control
\begin{equation}
\bar{\mathbf{x}}_{t+j}
=
\sum_{k=1}^{K} w_t^{(k)} \mathbf{x}^{(k)}_{t+j},
\qquad
\bar{\mathbf{u}}_{t+j}
=
\sum_{k=1}^{K} w_t^{(k)} \mathbf{u}^{(k)}_{t+j}.
\label{eq:nominal_xu}
\end{equation}
We then define the rollout deviations
\begin{equation}
\delta \mathbf{x}^{(k)}_{t+j}
=
\mathbf{x}^{(k)}_{t+j}-\bar{\mathbf{x}}_{t+j},
\qquad
\delta \mathbf{u}^{(k)}_{t+j}
=
\mathbf{u}^{(k)}_{t+j}-\bar{\mathbf{u}}_{t+j}.
\label{eq:deviation_xu}
\end{equation}
To remove possible weighted bias, we further center them as
\begin{equation}
\tilde{\mathbf{x}}^{(k)}_{t+j}
=
\delta \mathbf{x}^{(k)}_{t+j}
-
\sum_{r=1}^{K} w_t^{(r)} \delta \mathbf{x}^{(r)}_{t+j},
\label{eq:centered_x}
\end{equation}
\begin{equation}
\tilde{\mathbf{u}}^{(k)}_{t+j}
=
\delta \mathbf{u}^{(k)}_{t+j}
-
\sum_{r=1}^{K} w_t^{(r)} \delta \mathbf{u}^{(r)}_{t+j}.
\label{eq:centered_u}
\end{equation}

Leveraging the powerful concept of state-feedback control, we fit a local linear relation
\begin{equation}
\tilde{\mathbf{u}}^{(k)}_{t+j} \approx \mathbf{K}_{t+j} \tilde{\mathbf{x}}^{(k)}_{t+j}, \qquad k=1,\dots,K.
\end{equation}
via weighted least squares:
\begin{equation}
\mathbf{K}_{t+j}
=
\arg\min_{\mathbf{K}}
\sum_{k=1}^{K}
w_t^{(k)}
\left\|
\tilde{\mathbf{u}}^{(k)}_{t+j}
-
\mathbf{K}\tilde{\mathbf{x}}^{(k)}_{t+j}
\right\|^2 .
\label{eq:wls}
\end{equation}
The closed-form solution is
\begin{equation}
\Sigma_{xx,t+j}
=
\sum_{k=1}^{K}
w_t^{(k)}
\tilde{\mathbf{x}}^{(k)}_{t+j}
\tilde{\mathbf{x}}^{(k)\top}_{t+j},
\label{eq:sxx}
\end{equation}
\begin{equation}
\Sigma_{ux,t+j}
=
\sum_{k=1}^{K}
w_t^{(k)}
\tilde{\mathbf{u}}^{(k)}_{t+j}
\tilde{\mathbf{x}}^{(k)\top}_{t+j},
\label{eq:sux}
\end{equation}
\begin{equation}
\mathbf{K}_{t+j}
=
\Sigma_{ux,t+j}
\left(
\Sigma_{xx,t+j}+\alpha \mathbf{I}
\right)^{-1},
\label{eq:Kt}
\end{equation}
where $\alpha \ge 0$ is a regularization parameter.

The planner outputs a policy package consisting of the feedforward control sequence, the reference state trajectory, and the extracted local feedback gains. At each control update instant, the applied control is computed as
\begin{equation}
\mathbf{u}^{\mathrm{cmd}}_t
=
\mathbf{u}_{\mathrm{ff},t}
+
\mathbf{K}_t\big(\mathbf{x}_t-\bar{\mathbf{x}}_t\big),
\label{eq:closed_loop_exec}
\end{equation}
where $\mathbf{x}_t$ is the current measured state at the control update instant, $\bar{\mathbf{x}}_t$ denotes the corresponding reference state on the weighted nominal rollout trajectory, and $\mathbf{u}_{\mathrm{ff},t}$ denotes the feedforward control at time step $t$ obtained from the MPPI update. In this way, the feedforward term provides predictive guidance, while the feedback term compensates for delay-induced state mismatch and local execution errors at the moment of command application.

In practice, the effective delay of each planning cycle can be measured online as
\begin{equation}
\tau_t = s_t^{\mathrm{apply}} - s_t^{\mathrm{plan}},
\label{eq:tau_measure}
\end{equation}
where $s_t^{\mathrm{plan}}$ denotes the timestamp at which the planning cycle at iteration $t$ starts, and $s_t^{\mathrm{apply}}$ denotes the first control update instant at which the newly planned policy is actually applied. These delay samples are then used to estimate delay statistics, such as $\mathbb{E}[\tau]$, $\mathrm{Var}(\tau)$, and $\mathbb{E}[\tau^2]$, which in turn are used to adapt $\Sigma_{\mathrm{delay}}$.
To improve robustness under weight collapse, we additionally employ command saturation, rate limits, and an effective-sample-size threshold during execution.

CLIS-MPPI therefore transforms execution delay into an initial-state uncertainty distribution and augments MPPI with a rollout-based local feedback policy for closed-loop execution.

\section{EXPERIMENTS}
In this section, we validate the proposed CLIS-MPPI through real-world experiments on a 6-DoF manipulator for obstacle-avoidance and goal-reaching tasks. As we shall observe, compared with standard MPPI, the experimental results verify the effectiveness of the proposed method in narrow-space obstacle avoidance, achieving collision-free execution, shorter total path length, and significantly lower error relative to the reference state.

\subsection{Real-World Experimental Setup}

\begin{figure}[htbp]
    \centering
    \includegraphics[width=1.0\linewidth]{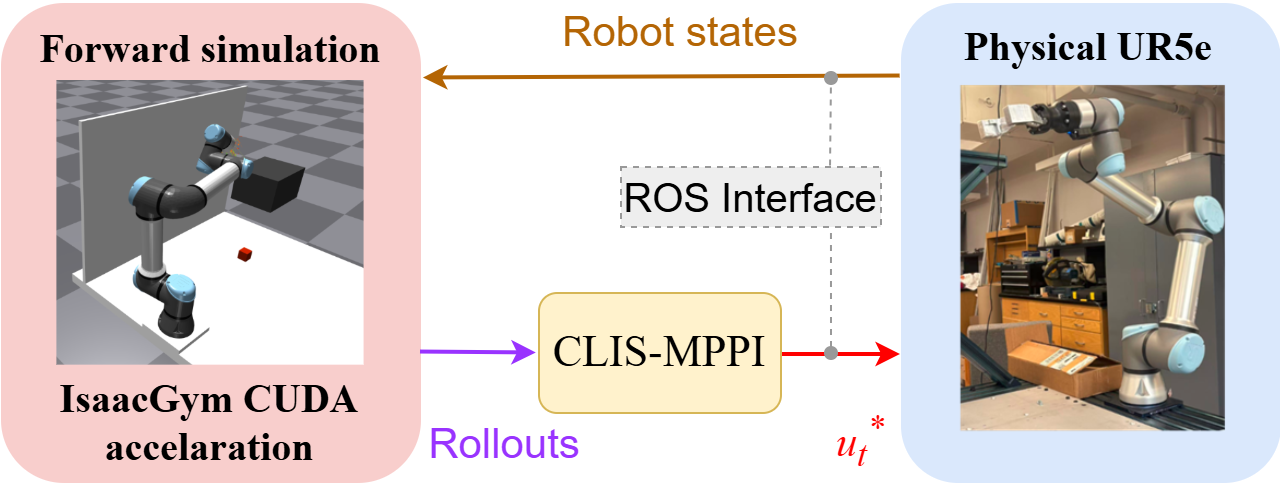}
    \caption{Experimental setup.}
    \label{experiment_setup}
\end{figure}

\subsubsection{Hardware and implementation}
The experimental platform is shown in Fig.~\ref{experiment_setup}. We conduct experiments on a 6-DoF UR5e manipulator. A PC equipped with an Intel Core i7-9750H CPU, an NVIDIA GeForce GTX 1650 GPU, and 32\,GB RAM serves as the onboard computing platform. The forward simulation of MPPI rollouts is performed in Isaac Gym, leveraging CUDA-based parallel computation to improve runtime efficiency.

\subsubsection{State and control interface}
We read the UR5e joint states via ROS, using a 12-dimensional state vector
$\mathbf{x}_t=[\mathbf{q}_t^\top,\dot{\mathbf{q}}_t^\top]^\top\in\mathbb{R}^{12}$,
where $\mathbf{q}_t\in\mathbb{R}^{6}$ and $\dot{\mathbf{q}}_t\in\mathbb{R}^{6}$ denote joint positions and velocities. CLIS-MPPI outputs a joint-velocity command $\mathbf{u}_t\in\mathbb{R}^{6}$, which is sent back to the UR5e through ROS for receding-horizon execution. Due to GPU memory constraints, we set the control rate to 10\,Hz ($dt=0.1$\,s). The other key parameters of the real-world experiment are summarized in Table~\ref{tab:real-params}.

\begin{table}[htbp]
\centering
\caption{Real-world experiment parameters}
\label{tab:real-params}
\setlength{\tabcolsep}{5pt}
\renewcommand{\arraystretch}{1.05}
\begin{tabular}{l c}
\hline
\textbf{Parameter} & \textbf{Value} \\
\hline
Rollouts $K$ & 50 \\
Horizon $N$ & 12 \\
Control step $dt$ & 0.1 s (10 Hz) \\
Joint velocity bound $|\dot{q}_i|$ & $\le 0.2$ rad/s \\
Joint position bound $|q_i|$ & $\le 2\pi$ rad \\
Temperature $\lambda$ & 0.05 \\
\hline
\end{tabular}
\vspace{-0.6em}
\end{table}

\subsubsection{Cost function}
We use a goal-reaching objective with collision penalties. Let $\mathbf{p}_{ee}(\mathbf{q}_t)\in\mathbb{R}^3$ and $R_{ee}(\mathbf{q}_t)\in SO(3)$ denote the end-effector position and orientation from forward kinematics, and let $(\mathbf{p}_g,R_g)$ be the goal pose. The stage cost is defined as
\begin{equation}
\begin{aligned}
\ell(\mathbf{x}_t,\mathbf{u}_t)
&= w_p\left\|\mathbf{p}_{ee}(\mathbf{q}_t)-\mathbf{p}_g\right\|_2
+ w_R\, d_R\!\big(R_{ee}(\mathbf{q}_t),R_g\big) \\
&\quad + w_c\, \mathbb{I}_{\mathrm{col}}(\mathbf{q}_t),
\end{aligned}
\end{equation}
where $d_R(\cdot,\cdot)$ measures orientation mismatch (e.g., quaternion angle error), and $\mathbb{I}_{\mathrm{col}}(\mathbf{q}_t)\in\{0,1\}$ indicates collision or contact with the table, shelf and obstacle. For the UR5e robot and our task space, we set $(w_p,w_R,w_c)=(10,3,1000)$.

\subsection{Experimental Results}

\begin{figure*}[t]
    \centering
    \includegraphics[width=1\textwidth]{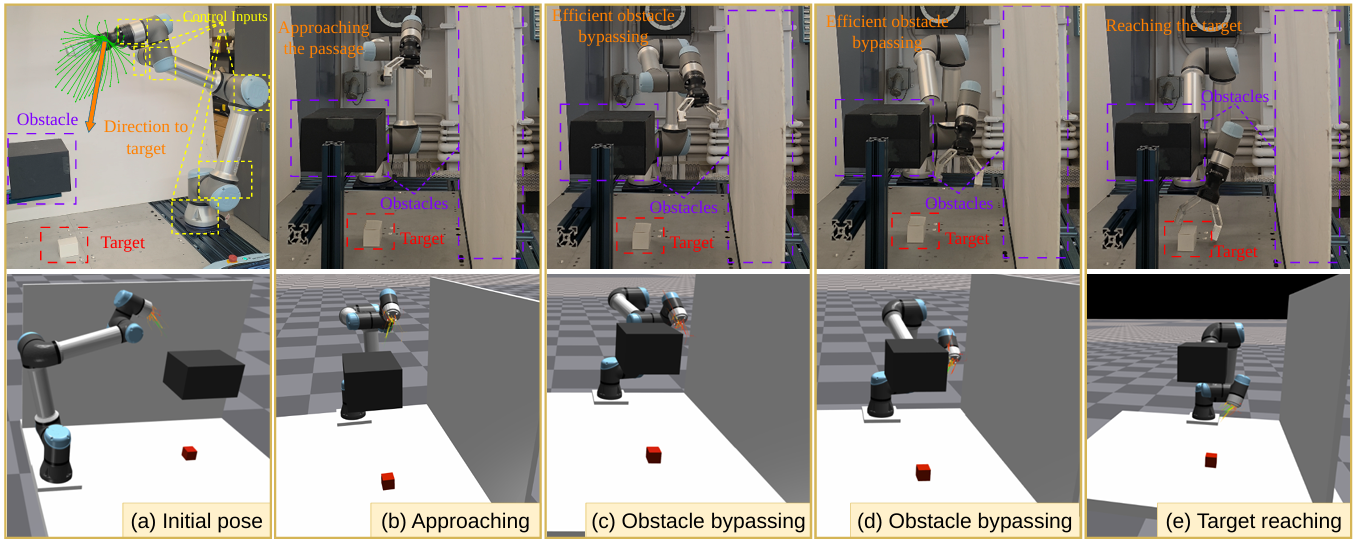}
    \caption{Real-world UR5e tabletop experiment. A black box serves as the obstacle and the cube on the table is the end-effector goal target.}
    \label{experiments}
\end{figure*}

The experimental task and representative execution sequence are shown in Fig.~\ref{experiments}. The scene is set on a tabletop with a black box placed in front of the robot as an obstacle. A cube on the table serves as the goal target for the end effector to reach. In particular, the gap between the black obstacle and the vertical wall is narrow, leaving only a limited collision-free passage for the manipulator to move through. This setting makes the task highly sensitive to execution-time state mismatch and therefore suitable for evaluating the proposed method.



\begin{table}[htbp]
\centering
\caption{Real-world results on the UR5e tabletop setup.}
\label{tab:real-results}
\setlength{\tabcolsep}{6pt}
\renewcommand{\arraystretch}{1.05}
\begin{tabular}{l c c}
\toprule
\textbf{Method} & MPPI & \textbf{CLIS-MPPI} \\
\midrule
\textbf{Collision-free}    & No & \textbf{Yes} \\
\textbf{Mean Planning Time (ms)}    & 87.5 & 87.6 \\
\textbf{Task Completion Time (s)}    & 25.4 & \textbf{24.4} \\
\textbf{End-effector Path (m)}    & 0.456 & \textbf{0.425} \\
\bottomrule
\end{tabular}
\vspace{-0.6em}
\end{table}


    
    
    

\begin{figure}[htbp]
    \centering
    \includegraphics[width=1.0\linewidth]{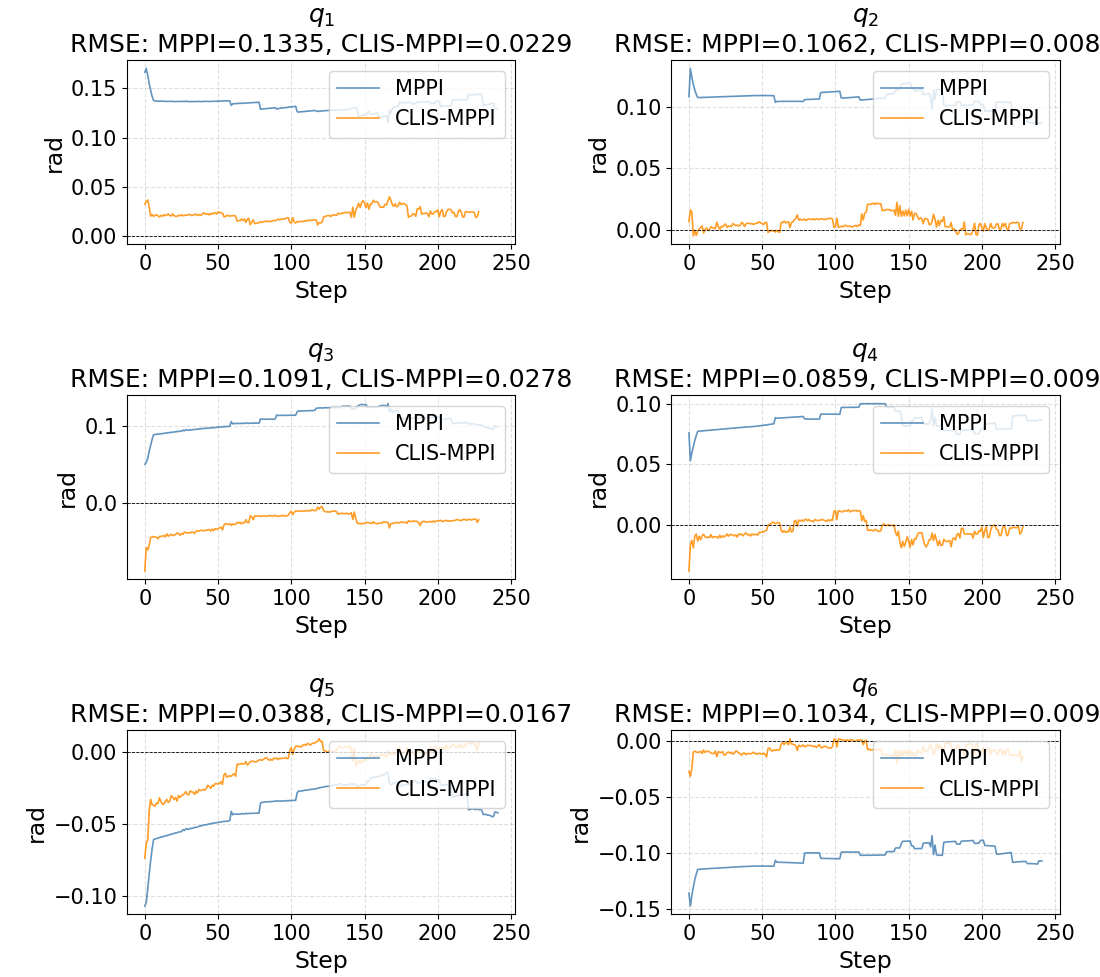}
    \caption{Joint position errors with respect to the forward-simulated reference joint positions for MPPI and CLIS-MPPI}
    \label{pos_errors}
\end{figure}

\begin{figure}[htbp]
    \centering
    \includegraphics[width=1.0\linewidth]{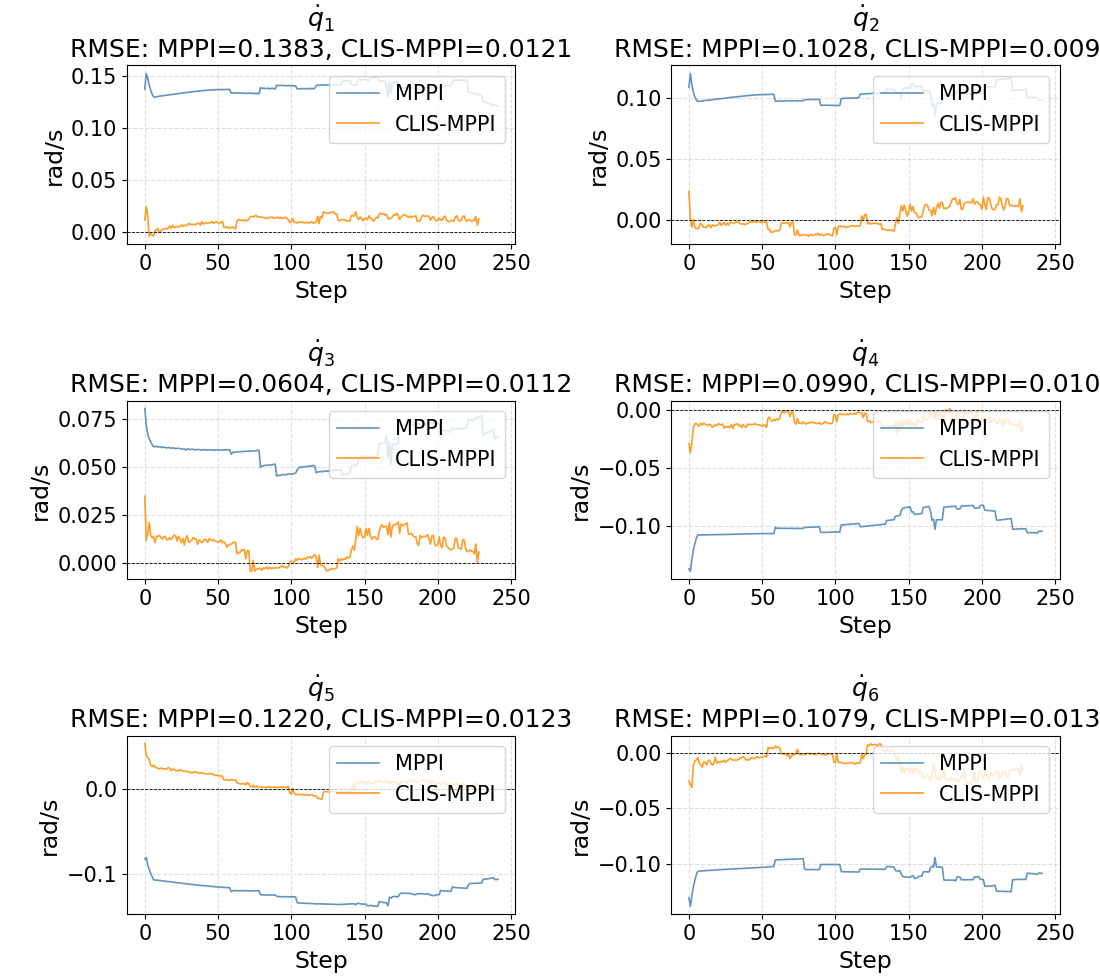}
    \caption{Joint velocity errors with respect to the forward-simulated reference joint velocities\ for MPPI and CLIS-MPPI}
    \label{vel_errors}
\end{figure}

The experimental results of MPPI and CLIS-MPPI are summarized in Table~\ref{tab:real-results} and Figs.~\ref{experiments}--\ref{vel_errors}. 
A key observation is that the baseline MPPI is not consistently collision-free in this real-world setting. 
Although the control commands planned by MPPI are typically collision-avoiding in forward simulation, collisions with the vertical wall still occur during execution. 
Specifically, in 5 repeated trials, MPPI collided in 2 runs. 
This discrepancy is mainly caused by execution-time state mismatch. 
As shown in Table~\ref{tab:real-results}, the average planning time of MPPI is 87.5 ms per control cycle. 
During this computation interval, the robot continues executing the control command from the previous cycle, while its actual state keeps evolving. 
As a result, by the time the newly planned command is applied, the robot state may have already deviated significantly from the state at which the current planning cycle started. 
Such mismatch becomes especially critical in the narrow passage shown in Fig.~\ref{experiments}, where even a small deviation can lead to contact with the obstacle or the wall.

In contrast, CLIS-MPPI completely avoids this issue in all real-world trials. 
The proposed method explicitly incorporates delay-aware initial-state sampling during forward simulation and generates closed-loop control inputs with online state-feedback correction. 
Therefore, the applied control command is no longer based solely on an outdated state at the beginning of the planning cycle, but instead continuously compensates for execution delay and state deviation in an online manner. 
This enables the robot to more accurately track the predicted system evolution and maintain safe motion through the narrow collision-free passage.

As reported in Table~\ref{tab:real-results}, the task completion time and end-effector path length are averaged over all collision-free trials for each method, corresponding to 3 trials for MPPI and 5 trials for CLIS-MPPI. Among their respective collision-free trials, CLIS-MPPI achieves better performance, reducing the average task completion time from 25.4~s to 24.4~s and shortening the average end-effector path length from 0.456~m to 0.425~m.
This improvement indicates that CLIS-MPPI not only enhances safety, but also improves motion efficiency. 
The reason is that, with more accurate state estimation under execution delay, each control command can be applied to a state that more closely matches the actual robot configuration, leading to more precise and effective motion generation.

Figs.~\ref{pos_errors} and~\ref{vel_errors} further demonstrate the state-consistency advantage of CLIS-MPPI over MPPI. 
The error curves shown in these figures are taken from the trial corresponding to the median end-effector path length among the 3 valid collision-free trials.
For all joints, CLIS-MPPI yields substantially smaller errors with respect to both the reference joint positions and the reference joint velocities. 
The corresponding RMSE values are consistently much lower than those of MPPI, often by a factor of several. In particular, the overall joint position and joint velocity RMSEs of MPPI are 0.100 rad and 0.108 rad/s, respectively, whereas those of CLIS-MPPI are 0.017 rad and 0.011 rad/s, representing a 5.9- and a 9.7-fold reduction, respectively.
These results confirm that CLIS-MPPI enables the actual robot state to remain much closer to the forward-simulated reference state at each control step throughout execution. 
As a consequence, the robot motion becomes smoother and more stable, with less noticeable hesitation or jerky behavior during obstacle avoidance and target reaching.

\section{CONCLUSIONS}
This paper proposed CLIS-MPPI, a delay-aware closed-loop MPPI for real-world robotic systems. By explicitly modeling planning delay as initial-state uncertainty and combining MPPI with online state-feedback correction, the proposed method substantially mitigates the mismatch between the state used in forward simulation and the actual state at which the control command is applied. As a result, CLIS-MPPI effectively improves the safety of MPPI in delay-sensitive tasks, for example by preventing collisions that can arise from delay-induced state mismatch, while also producing smoother and more stable robot motion during execution. The effectiveness of the proposed method was validated through real-world experiments on a 6-DoF UR5e manipulator, where CLIS-MPPI demonstrated improved collision avoidance, state consistency, and motion smoothness compared with standard MPPI.

\addtolength{\textheight}{-12cm}   





\section*{ACKNOWLEDGMENT}
Support for this research was provided in part by an Amazon Science-Hub gift award. The views and conclusions contained in this document are those of the authors and should not be interpreted as representing the official policies, either expressed or implied, of the sponsoring organization or the University of Washington.




\bibliographystyle{IEEEtran}
\bibliography{CDC2026Ref_Liuzd}


\end{document}